\pdfoutput=1
\documentclass{spie}

\usepackage{cite}

\usepackage{times} 
\usepackage{indentfirst} 

\usepackage{amsmath} 
\usepackage{tikz}
\usepackage{multirow}
\usepackage{multicol}

\usepackage[utf8x]{inputenc}
\usepackage[english]{babel}

\usepackage[colorinlistoftodos,color=pink]{todonotes} 
\presetkeys{todonotes}{inline}{}

\newcommand{\relu}{\mbox{ReLU}}
\newcommand{\morph}{\mbox{BMN}}

\usepackage[ruled,linesnumbered]{algorithm2e}
\usetikzlibrary{automata, positioning}
\usepackage{mathtools,amssymb}

\title{Bipolar Morphological Neural Networks: Convolution Without Multiplication}

\author{Elena Limonova\supit{1,2,4}, Daniil Matveev\supit{2,3}, Dmitry Nikolaev\supit{2,4}, Vladimir V. Arlazarov\supit{2,5}
  \skiplinehalf
  \normalsize 
  \supit{1} Institute for Systems Analysis, FRC CSC RAS, Moscow, Russia; \\
  \supit{2} Smart Engines Service LLC, Moscow, Russia;\\
  \supit{3} Moscow State University, Moscow, Russia;\\
  \supit{4} Institute for Information Transmission Problems RAS, Moscow, Russia; \\
  \supit{5} Moscow Institute of Physics and Technology, Dolgoprudnii, Russia;
}

\begin{document}

\maketitle

\begin{abstract}
In the paper we introduce a novel bipolar morphological neuron and bipolar morphological layer models.
The models use only such operations as addition, subtraction and maximum inside the neuron and exponent and logarithm as activation functions for the layer. The proposed models unlike previously introduced morphological neural networks approximate the classical computations and show better recognition results.
We also propose layer-by-layer approach to train the bipolar morphological networks, which can be further developed to an incremental approach for separate neurons to get higher accuracy. 
Both these approaches do not require special training algorithms and can use a variety of gradient descent methods.
To demonstrate efficiency of the proposed model we consider classical convolutional neural networks and convert the pre-trained convolutional layers to the bipolar morphological layers.
Seeing that the experiments on recognition of MNIST and MRZ symbols show only moderate decrease of accuracy after conversion and training, bipolar neuron model can provide faster inference and be very useful in mobile and embedded systems.

\keywords{morphological neural network, bipolar neuron, computational efficiency, layer-by-layer training}
\end{abstract}

\section{Introduction}

Neural network methods are widely used in problems of recognition and machine vision \cite{skoryukina2015real, bulatov2018smart, abramov2016, avsentiev2017, sheshkus2015approach}. Various deep neural network architectures have been developed for solving problems of current interest. The scope of neural network usage is steadily growing. Currently they are actively used on mobile devices and embedded systems with limited performance and a strong need for low power consumption. Despite the fact that there is a number of methods for improving inference speed of neural networks \cite {Gupta, Denton, ilin2017fast, limonova2018convolutional}, each of them has its own constraints that limit applicability, so this area is still the area of active research.

One way to improve the inference time of neural networks is to use a computationally simplified neuron model. The calculations in such a neuron model can be implemented using fewer logic gates than the sequence of multiplications and additions used in the classical neuron model. This means that calculations in a simplified neuron can be performed in less time and are more energy efficient. The latter circumstance is especially important for mobile recognition systems.

The examples of neural networks with simplified models of a neuron are neural networks with integer calculations \cite{google} and morphological neural networks \cite {Ritter1996}. The usage of integer data types can speed up inference because calculation of the integer sum and integer product on modern mobile central processors is faster than real sum and product. This is due to the architectural issues of the ARM processor, which is most often used on mobile devices and embedded systems, as well as the presence of Single Instruction Multiple Data extensions. Such extensions can perform one operation on the elements of data register simultaneously \cite{Patterson}. In case of integer calculations it is very useful, as the register has fixed size of, for example, 128 bits (for NEON and SSE) and allows processing of 4 float32 values and 16 8-bit values. However, replacing the classical neuron model with an integer model implies a change in the calculation results due to accuracy loss of the weights and possible overflows. Recent research introduces different methods to preserve recognition quality even with low-bit quantization of a network \cite{Gupta, zhou2016dorefa, courbariaux2016binarized, rastegari2016xnor, ilin2017fast, Zhou2017, Choukroun2019}.

The morphological neuron model uses addition and maximum/minimum operations instead of addition and multiplication, as well as threshold nonlinear activation functions \cite {Ritter1996, Sussner2009}. This model largely appeals to the biological properties of neurons. Further development of this idea is a dendrite morphological neuron, which allows simulating the processes of excitation and inhibition \cite {Ritter2003_dendrites} and a generalization of the proposed model in terms of lattice algebra \cite {Ritter2003}. It is also worth noting that the morphological neural network is usually a single-layer perceptron. To train such a neural network, heuristic algorithms are used \cite{Socca2014}, which can be supplemented by stochastic gradient descent \cite {Zamora2016}.

One more morphological neural network model DenMo-Net with dilation and erosion neurons was presented in \cite{mondal2019dense}. The network demonstrated good results with three-layer DenMo-Net architecture relative to a classical three-layer architecture. However, the considered model does not seem to be scalable and does not show state-of-the-art quality in image recognition problems with such a simple structure.

In this paper, a new model of neuron based on the idea of a morphological neuron is proposed. The proposed model is considered as an approximation of a classical neuron, which allows us to adapt modern neural network architectures to this model. We also introduce an approach to training and fine-tuning of such neural networks. We tested it for the MNIST number recognition \cite{MNIST} and Machine-Readable Zone (MRZ) \cite{mrz} symbol recognition problems. Experimental results show no recognition quality loss in the problems.

\section{Bipolar morphological layer model}

We propose approximation of neural network layers with neurons calculating linear combination of inputs by a morphological structure. Such layers can be fully-connected or convolutional, which are normally the most computationally complex part of a neural network. Let us describe the proposed morphological structure. We call it bipolar morphological layer (BM layer) or bipolar morphological neuron (BMN) for one neuron. The word bipolar here refers to the two computational paths, that consider excitation and inhibition processes in the neuron. In biology bipolar neurons are mostly responsible for perception and can be found, for example, in the retina.
Our main idea is to represent the major amount of computations using max/min and addition operations.

The structure of BM layer is inspired by approximation of the classical layer. Calculations of one neuron can be expressed as 4 neurons placed in parallel with the following addition/subtraction:
$$
\sum_{i=1}^N x_i w_i = \sum_{i=1}^N p_i^{00} x_i w_i - \sum_{i=1}^N p_i^{01} x_i |w_i| - \sum_{i=1}^N p_i^{10} |x_i| w_i + \sum_{i=1}^N p_i^{11} |x_i| |w_i|,
$$
where
$$
p_i^{kj} =
    \begin{cases}
    1 , \mbox{ if } (-1)^k x_i > 0 \mbox{ and } (-1)^j w_i > 0\\
    0 , \mbox{ otherwise}
    \end{cases}
$$

Values of $p_i^{kj}$ define the connections of new neurons.
For each of them we can consider inputs and weights as positive values and perform the approximation. Let us denote:
\begin{gather*}
M = \max_j (x_j w_j) \\
k = \frac{\sum \limits_{i=1}^N x_i w_i}{M} -1
\end{gather*}

The approximation is:
\begin{gather*}
\sum_{i=1}^N x_i w_i = \exp \lbrace \ln \sum_{i=1}^N x_i w_i \rbrace = \exp \left \{ \ln M (1 + k) \right \} = (1 + k)\exp \ln M = (1 + k) \exp \left \{ \ln(\max_j (x_j w_j))\right \} = \\
= (1 + k)\exp \max_j \ln (x_j w_j) = (1 + k)\exp \max_j (\ln x_j + \ln w_j)  = (1 + k)\exp \max_j (y_j + v_j) \approx \exp \max_j (y_j + v_j),
\end{gather*}
where $y_j$ are new inputs, $v_j = \ln w_j$ are new weights. It is obvious, that it is correct, when $k \ll 1$. Since $0 \leq k \leq N - 1$, the best case is the sum containing only one non-zero term ($k = 0$) and the worst case is the sum with all equal terms ($k = N-1$). In the worst case real value for the sum will be $N$ times more than approximated. This behavior can not be called a good approximation. However, even an approximation with well limited absolute error can lead to an unpredictable decrease of neural network accuracy due to a strong non-linearity between layers. For example, low-precision neural networks do not show high accuracy after direct conversion and give perfect results with the help of special training approaches like in \cite{Zhou2017}. Thus, the accuracy of the one neuron approximation should not be a decisive criterion in the case of neural networks. It is much more important whether the approximation leads to a high resulting quality of the network or not. In the paper we investigate the accuracy of the introduced approximation after conversion and training.

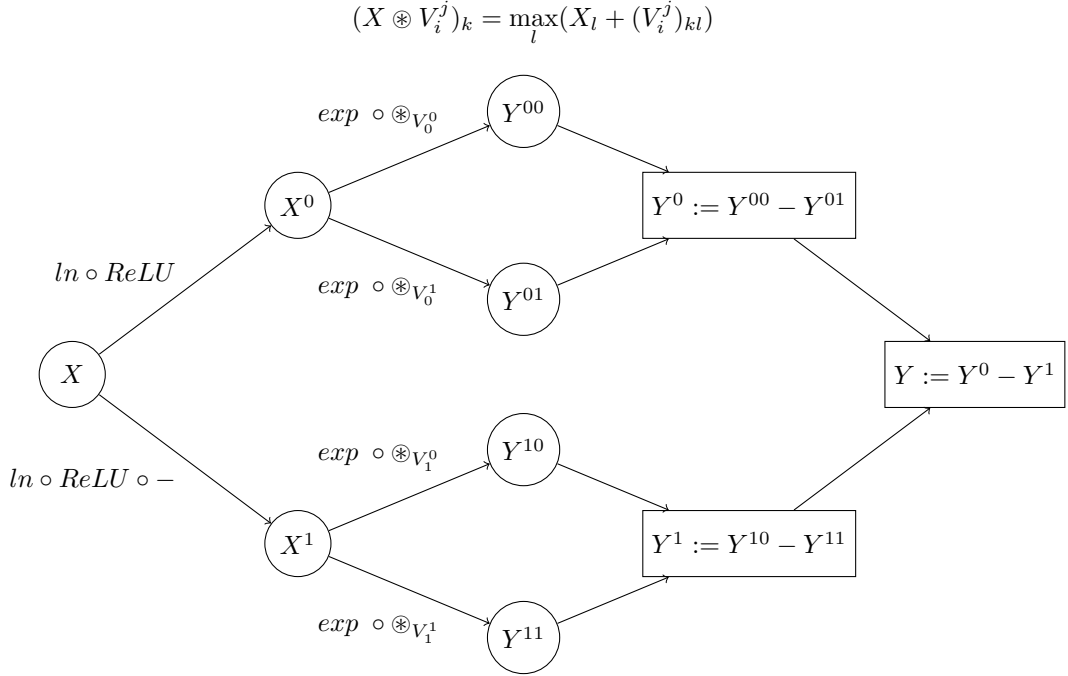
\begin{figure}[h]
\centering
$$
(X \circledast V_i^j)_k = \max_l (X_l + (V_i^j)_{kl})
$$
\begin{tikzpicture}[node distance=2cm,auto]
\node[state] (a) at (0,0) {$X$};
\node[state] (b) at (3,2.25) {$X^0$};
\node[state] (c) at (3,-2.25) {$X^1$};
\node[state] (d) at (6,3.5) {$Y^{00}$};
\node[state] (e) at (6,1) {$Y^{01}$};
\node[state] (f) at (6,-1) {$Y^{10}$};
\node[state] (g) at (6,-3.5) {$Y^{11}$};
\node[state] (h) at (9,2.25) [rectangle,draw] {$Y^0:=Y^{00}-Y^{01}$};
\node[state] (i) at (9,-2.25) [rectangle,draw] {$Y^1:=Y^{10}-Y^{11}$};
\node[state] (j) at (12,0) [rectangle,draw] {$Y:=Y^0-Y^1$};

\path[->] (a) edge node {$\large {ln \circ ReLU}$} (b);
\path[->] (a) edge node[swap] {$ln \circ ReLU \circ -$} (c);
\path[->] (b) edge node [near end] {$exp \; \circ \circledast_{V_0^0}$} (d);
\path[->] (b) edge [near end] node[swap] {$exp \; \circ \circledast_{V_0^1}$} (e);
\path[->] (d) edge node {$ $} (h);
\path[->] (e) edge node {$ $} (h);
\path[->] (c) edge [near end] node {$exp \; \circ \circledast_{V_1^0}$} (f);
\path[->] (c) edge [near end] node[swap] {$exp \; \circ \circledast_{V_1^1}$} (g);
\path[->] (f) edge node {$ $} (i);
\path[->] (g) edge node {$ $} (i);
\path[->] (h) edge node {$ $} (j);
\path[->] (i) edge node {$ $} (j);
\end{tikzpicture}
\caption{The structure of a BM layer with input vector $X$ and weight matrix $V_i^j$. The $\circ$ symbol designates function composition.}
\label{structure}
\end{figure}

The approximation shown above leads to the proposed BM layer structure in the Fig. \ref{structure}. The rectifier (ReLU) allows us to take values above zero and create four computational path for positive and negative input or coefficients. Then we take the logarithm of rectified input ($X$) and perform essential morphological operation of the layer. The results are passed to the exponential unit and subtracted to get the output ($Y$).

The BM layer will obtain results similar to the original in case of a good logarithm approximation, which means that the sum has one major term. If there are several dominant terms we can take it into account for our approximation.

Operations of $\ln$ and $\exp$ are performed on the activation (the signal transmitted between network layers) and can be considered as a part of an activation function. Normally the activation functions do not make significant contributions to the computational complexity of a neural network, so the increase of computational complexity should be of little consequence. If the activation function yet take noticeable time, we can approximate it by a simpler function. For example, it can be a piece-wise linear approximation. One more option is to perform input quantization and use look-up tables, which is also fast.

As a result, BM layer structure can be expressed as follows:
\begin{gather*}
\morph(x, w) = \exp{\max_j (\ln \relu(x_j) + v^{0}_j)} - \exp{\max_j (\ln \relu(x_j) + v^{1}_j)} - \exp{\max_j (\ln \relu(-x_j) + v^{0}_j)} + \\
+ \exp{\max_j (\ln \relu(-x_j) + v^{1}_j)},
\end{gather*}
where 
$$
v_j^{k} =
    \begin{cases}
    \ln |w_j| , \mbox{ if } (-1)^k w_j > 0\\
    -\infty , \mbox{ otherwise}
    \end{cases}
$$

If a layer of the neural network includes bias, which is added to the linear combination, it can be added after the proposed BM layer approximation.

\section{Training method}

Let us introduce a method for obtaining a neural network with BM layers. Training BM using standard algorithms can be challenging, because there is only one non-zero gradient element due to max operation and only one weight is updated at each iteration. Some weights can never be updated and never fire after training thus giving redundancy of the network. 
We use their approximation nature and train them layer-by-layer. The idea is to modify convolutional and fully-connected layers sequentially from the first to the last, freeze the modified structure and weights and train the other layers. The approach allows us to fine-tune the network and ignore possible issues of training the BMNs directly and still adapt the neural network to the changes and save calculation accuracy. We have obtained good results by this layer-by-layer method while training 8-bit integer neural networks \cite{ilin2017fast}. The idea to divide weights into groups and perform approximation and fine-tuning until the approximation of the full network is introduced in \cite{Zhou2017} for lossless low-bit quantization. 

As a result, the method 1 of training can be summarized as:

\begin{algorithm}[H]
\caption{Training of BM network}
\KwData{Training data}
\KwResult{Neural network with BM layers}
Train classical neural network by standard methods\;
\lForEach{conv and fc layers}{\\
  \Indp Approximate current layer and freeze its weights;\\
  Train the remaining part of the network by standard methods}
\Indm Perform steps 1-4 several times with different initial conditions and choose the best result\;
\end{algorithm}

We also present the second training method, which is different from method 1 at stages 3 and 4. The weights are not frozen and BM layers are trained with the whole network. In this case we can face convergence issues and slower training process, but try to avoid them by initialization with converted weights, that we suppose are close to the desired values.

The next step for the algorithm development is using neuron-by-neuron fine-tuning. Then we should perform approximation and freezing of weights in a layer only for one neuron at a time. 

\section{Experimental results}

\subsection{MNIST}

\begin{figure}[h!]
    \begin{minipage}[h]{0.4\linewidth}
    \center{\includegraphics[width=1\linewidth]{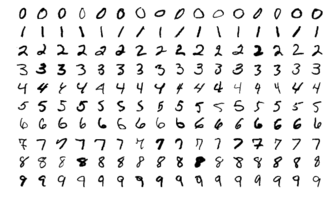} \\ a)}
    \end{minipage}
    \hfill
    \begin{minipage}[h]{0.4\linewidth}
    \center{\includegraphics[width=1\linewidth]{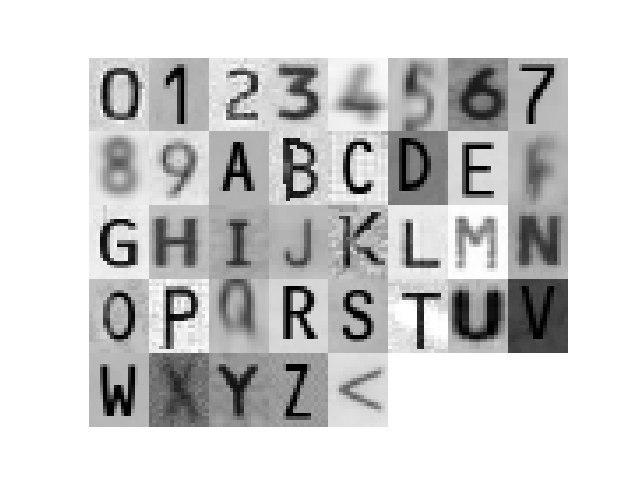} \\ b)}
    \end{minipage}
    \caption{The example images from training datasets. a) MNIST, b) MRZ}
    \label{fig_imgs}
\end{figure}

MNIST is a database with images of handwritten digits. This is a basic dataset for image classification problems. The training set consists of 60000 gray images of $28 \times 28$ pixels \cite{MNIST}. There is also a test data of 10000 images. We use 10\% of the training set for validation and the rest for training. The examples of MNIST images are shown in Fig. \ref{fig_imgs}a.

We consider 2 simple convolutional neural networks (CNNs) and analyze the accuracy obtained after replacing convolutional and fully-connected layers with BM layers.

The notation used to represent different architectures is as follows:

conv($n$, $w_x$, $w_y$) --- convolutional layer with $n$ filters of size $w_x \times w_y$;

fc($n$) --- fully-connected layer with $n$ neurons;

maxpool($w_x$, $w_y$) --- max-pooling layer with the window of size $w_x \times w_y$;

dropout($p$) --- dropout the input signals with the probability $p$;

relu --- rectifier activation function ${\mbox{ReLU}(x) = \max(x, 0)}$;

softmax --- standard softmax activation function.

The CNN$_1$ architecture is: conv1(30, 5, 5) - relu1 - dropout1(0,2) - fc1(10) - softmax1.

The CNN$_2$ architecture is: conv1(40, 5, 5) - relu1 - maxpool1(2, 2) - conv2(40, 5, 5) - relu2 - fc1(200) - relu3 - dropout1(0,3) - fc2(10) - softmax1.

For CNN$_1$ and CNN$_2$ we perform layer-by-layer conversion to the BM network. In the Table \ref{table_mnist} we show the resulting accuracy depending on the converted part. Converted part "none" corresponds to the original classical network. For methods 1 and 2 we demonstrate the accuracy before fine-tuning (BM layer's weights are not trained) and after fine-tuning (the whole network is trained). All values were averaged over 10 measurements with random initialization.

The results for training without BM layers show moderate accuracy decrease for convolutional layers and dramatic decline for fully-connected layers. This can happen due to the drop of the approximation quality for these layers. Moreover, accuracy with two convolution layers converted is not much better than the accuracy of the fully-connected layers only, which means that BM convolutions without training perform only slightly better than random. However, results with training of converted layers show almost no accuracy decrease after convolution layers conversion and better results for fully-connected layers.

Introduced results do not necessary mean that BM neural network with fully-connected layers can not reach desired accuracy of classical networks, because training methods for BM neurons are still to be investigated. In our case conversion works excellently for convolutional layers, but gives poor results for fully-connected ones. At the same time, neural network inference takes major time for convolutional layers, so the proposed method suits well for speeding up inference.

The area of future research is developing of new training approaches to achieve state-of-the-art quality for multiplication-free neural network.

\begin{table}[h!]
\centering
\caption{MNIST recognition accuracy for neural networks with BM layers.}
\label{table_mnist}
\begin{tabular}{| p{0.05\textwidth} | p{0.3\textwidth} | p{0.10\textwidth} | p{0.10\textwidth} | p{0.10\textwidth} | p{0.10\textwidth} |}
  \hline
  \multirow{3}{*}{CNN} & \multirow{3}{*}{Converted part} & \multicolumn{4}{c|}{Accuracy, \%} \\ \cline{3-6} 
  & & \multicolumn{2}{c|}{Method 1} & \multicolumn{2}{c|}{Method 2} \\ \cline{3-6} 
  & & before fine-tuning & after fine-tuning & before fine-tuning & after fine-tuning \\ \hline
  \multirow{3}{*}{CNN$_1$} & none & \textbf{98,72} & - & \textbf{98,72} & -\\ \cline{2-6}
   & conv1 & 42,47 & 98,51 & 38,38 & \textbf{98,76}\\ \cline{2-6}
   & conv1 - relu1 - dropout1 - fc1 & 26,89 & - & 19,86 & 94,00 \\ \hline
   \multirow{4}{*}{CNN$_2$} & none & \textbf{99,45} & - & \textbf{99,45} & -\\ \cline{2-6}
    & conv1 & 94,90 & 99,41 & 94,57 & 99,42\\ \cline{2-6}
    & conv1 - relu1 - maxpool1 - conv2 & 21,25 & 98,68 & 36,23 & \textbf{99,37}\\ \cline{2-6}
    & conv1 - relu1 - maxpool1 - conv2 - relu2 - fc1 & 10,01  & 74,95 & 17,25 & 99,04 \\ \cline{2-6}
    & conv1 - relu1 - maxpool1 - conv2 - relu2 - fc1 - dropout1 - relu3 - fc2 & 12,91  & - & 48,73 & 97,86\\
  \hline
\end{tabular}
\end{table}

\subsection{MRZ symbols}

Here we use private dataset of about $2,8 \times 10^5$ gray images of $21 \times 17$ pixels in size. The images contain 37 MRZ symbols, which were extracted from real documents with machine-readable zone. We use 10\% of the data for validation and $9,4 \times 10^4$ additional images for test.

The CNN$_3$ architecture is: conv1(8, 3, 3) - relu1 - conv2(30, 5, 5) - relu2 - conv3(30, 5, 5) - relu3 - dropout1(0,25) - fc1(37) - softmax1.

The CNN$_4$ architecture is: conv1(8, 3, 3) - relu1 - conv2(8, 5, 5) - relu2 - conv3(8, 3, 3) - relu3 - dropout1(0,25) - conv4(12, 5, 5) - relu4 - conv5(12, 3, 3) - relu5 - conv6(12, 1, 1) - relu6 - fc1(37) - softmax1.

Conversion results are shown in the Table \ref{table_mrz}.
Converted part ``one`` corresponds to the original classical network. For methods 1 and 2 we demonstrate the accuracy before fine-tuning (BM layer's weights are not trained) and after fine-tuning (the whole network is trained).
All values were averaged over 10 measurements with random initialization.

The recognition accuracy is only slightly different for the first two converted layers and trained remaining part of the network (with frozen BM layers), but then significantly decreases. The possible reason of the effect can be difficulty with adapting to new approximate features extracted by the BM convolutional layers. However, training of the full converted network including BM layers shows no significant accuracy decline for all convolutional layers, but shows visible degradation for fully-connected layers. 
As a result, currently we recommend conversion and training of convolutional layers only to keep the original recognition quality, while BM neurons for fully-connected layers are to be further investigated. 

\begin{table}[h!]
\centering
\caption{MRZ recognition accuracy for neural networks with BM layers.}
\label{table_mrz}
\begin{tabular}{| p{0.05\textwidth} | p{0.3\textwidth} | p{0.10\textwidth} | p{0.10\textwidth} | p{0.10\textwidth} | p{0.10\textwidth} |}
  \hline
  \multirow{3}{*}{CNN} & \multirow{3}{*}{Converted part} & \multicolumn{4}{c|}{Accuracy, \%} \\ \cline{3-6} 
  & & \multicolumn{2}{c|}{Method 1} & \multicolumn{2}{c|}{Method 2} \\ \cline{3-6} 
  & & before fine-tuning & after fine-tuning & before fine-tuning & after fine-tuning \\ \hline
 \multirow{5}{*}{CNN$_3$} & none & \textbf{99,63} & - & \textbf{99,63} & - \\ \cline{2-6}
    & conv1 & 97,76 & 99,64 & 83,07 & 99,62\\ \cline{2-6}
    & conv1 - relu1 - conv2 & 8,59 & 99,47 & 21,12 & 99,58\\ \cline{2-6}
    & conv1 - relu1 - conv2 - relu2 - conv3 & 3,67 & 98,79 & 36,89 & \textbf{99,57}\\ \cline{2-6}
    & conv1 - relu1 - conv2 - relu2 - conv3 - relu3 - dropout1 - fc1 & 12,58  & - & 27,84 & 93,38\\
  \hline
   \multirow{8}{*}{CNN$_4$} & none & \textbf{99,67} & - & \textbf{99,67} & -\\ \cline{2-6}
    & conv1 & 91,20 & 99,66 & 93,71 & 99,67 \\ \cline{2-6}
    & conv1 - relu1 - conv2 & 6,14 & 99,52 & 73,79 & 99,66 \\ \cline{2-6}
    & conv1 - relu1 - conv2 - relu2 - conv3 & 23,58 & 99,42 & 70,25 & 99,66 \\ \cline{2-6}
    & conv1 - relu1 - conv2 - relu2 - conv3 - relu3 - dropout1 - conv4 & 29,56 & 99,04 & 77,92 & 99,63\\ \cline{2-6}
    & conv1 - relu1 - conv2 - relu2 - conv3 - relu3 - dropout1 - conv4 - relu4 - conv5 & 34,18 & 98,45 & 17,08 & 99,64 \\ \cline{2-6}
    & conv1 - relu1 - conv2 - relu2 - conv3 - relu3 - dropout1 - conv4 - relu4 - conv5 - relu5 - conv6& 5,83 & 98,00 & 90,46 & \textbf{99,61} \\ \cline{2-6}
    & conv1 - relu1 - conv2 - relu2 - conv3 - relu3 - dropout1 - conv4 - relu4 - conv5 - relu5 - conv6 -relu6 - fc1 & 4,70  & - & 27,57 & 95,46 \\
  \hline
\end{tabular}
\end{table}

\section{Conclusion and Discussion}

In the paper we propose a new bipolar morphological neuron, which approximates classical neuron. We show how to convert a layer of a classical neural network to the BM layer and introduce an approach to training. It utilizes layer-by-layer conversion to BM layers and training of the remaining part of the full network using standard methods. In such way the approach allows us to avoid training issues, such as updating only one weight at each step due to maximum operations. We demonstrate that recognition accuracy of BM networks with only convolutional layers converted is close to those of the original classical networks for MNIST and MRZ datasets.

Bipolar morphological neural networks give new possibilities to speed up neural network inference, because addition/subtraction and maximum/minimum have lower latency than multiplication for most modern devices. FPGA systems for BM neural networks can be easier and more energy efficient, because they do not require multiplication units for convolutions. In the proposed BM model we used complex activation functions, but they take much less time than convolutional or fully-connected layer, because they are applied to the activation signal between layers and have only linear complexity. Furthermore, activation functions can be approximated and implemented via look-up tables and be computed even faster.

It should be noted that state-of-the-art methods for speeding up inference like low-precision computations, pruning, or structure simplifications can also be applied to the BM model. However, the accuracy of the resulting neural networks is still to be determined.

One more advantage of the structure is that BM layers can be included in already existing architectures and do not restrict them in any aspects. For example, morphological neural networks do not allow us to stack many layers to increase quality, while here we can vary the number of BM layers without training concerns.

This work opens a way for the wide research of bipolar morphological neural networks as a method, which gives recognition accuracy close to those of classical neural networks (especially if converting only computationally complex parts) but better inference speed. Introduced training method can be further developed to allow, for example, training from scratch and improve results for BM fully-connected layers.

\acknowledgements

The reported study was partially supported by RFBR, research projects 17-29-03240 and 18-07-01384.

\bibliographystyle{spiebib_custom}
\bibliography{bibliography}

\end{document}